\documentclass{article}

\PassOptionsToPackage{numbers}{natbib}
\usepackage[preprint]{neurips_2024}

\usepackage[utf8]{inputenc} % allow utf-8 input
\usepackage[T1]{fontenc}    % use 8-bit T1 fonts
\usepackage{hyperref}       % hyperlinks
\usepackage{url}            % simple URL typesetting
\usepackage{booktabs}       % professional-quality tables
\usepackage{amsfonts}       % blackboard math symbols
\usepackage{nicefrac}       % compact symbols for 1/2, etc.
\usepackage{microtype}      % microtypography
\usepackage{xcolor}         % colors
\usepackage{graphicx}
\usepackage{subcaption}
\usepackage{array}
\usepackage{multirow}
\usepackage{xcolor} % Required for coloring text
\usepackage{wrapfig}
\usepackage{svg}
\usepackage{comment}
\usepackage{lineno} % Include the lineno package
\usepackage{amsmath}
\usepackage{amssymb}
\usepackage{graphicx}
\usepackage{overpic}
\usepackage{subcaption}
\graphicspath{{assets/figure-sources/}{assets/figures/}}

\definecolor{darkblue}{rgb}{0,0,0.75}

\hypersetup{
    hidelinks,  % Removes colored boxes from all links
    colorlinks=true,  % Optional: colors the text instead of boxing it
    linkcolor=red,  % Optional: sets the color of internal links
    citecolor=darkblue   % Optional: sets the color of citation links
}

\title{\textit{LTX-2}: Efficient Joint Audio-Visual Foundation Model}

\newcommand{\authsep}{\hspace{0.9em}}
\author{
\normalfont
Yoav HaCohen\thanks{Authors are listed with project leads first, followed by the team in alphabetical order.}\authsep
Benny Brazowski\authsep
Nisan Chiprut\authsep
Yaki Bitterman
\\
Andrew Kvochko\authsep
Avishai Berkowitz\authsep
Daniel Shalem\authsep
Daphna Lifschitz\authsep
Dudu Moshe\authsep
\\
Eitan Porat\authsep
Eitan Richardson\authsep
Guy Shiran\authsep
Itay Chachy\authsep
Jonathan Chetboun\authsep
\\
Michael Finkelson\authsep
Michael Kupchick\authsep
Nir Zabari\authsep
Nitzan Guetta\authsep
Noa Kotler\authsep
\\
Ofir Bibi\authsep
Ori Gordon\authsep
Poriya Panet\authsep
Roi Benita\authsep
Shahar Armon\authsep
\\
Victor Kulikov\authsep
Yaron Inger\authsep
Yonatan Shiftan\authsep
Zeev Melumian\authsep
Zeev Farbman
\\[0.5ex]
\textbf{Lightricks}\\
\texttt{ltx-2@lightricks.com}
}

\definecolor{darkgreen}{rgb}{0.0, 0.4, 0.0}

\defcitealias{ho2020denoising}{DDPM}
\defcitealias{brooks2024video}{Sora}
\defcitealias{chen2023pixart}{Pixart-$\alpha$}
\defcitealias{xie2024sana}{SANA}
\defcitealias{blattmann2023stable}{SVD}
\defcitealias{guo2023animatediff}{AnimateDiff}
\defcitealias{bar2024lumiere}{Lumiere}
\defcitealias{girdhar2024factorizing}{EmuVideo}
\defcitealias{polyak2024movie}{MovieGen}
\defcitealias{yang2024cogvideox}{CogVideoX}
\defcitealias{opensoraplan}{Open-Sora Plan}
\defcitealias{opensora}{Open-Sora}
\defcitealias{jin2024pyramidal}{PyramidFlow}
\defcitealias{chen2024deepcompressionautoencoderefficient}{DC-VAE}
\defcitealias{peebles2023scalable}{DiT}
\defcitealias{lu2024fit}{FiT}
\defcitealias{largedit}{LargeDiT}
\defcitealias{hoogeboom2023simple}{SimpleDiffusion}
\defcitealias{crowson2024scalable}{H-DiT}
\defcitealias{esser2024scaling}{SD3}
\defcitealias{chen2022scaling}{ScalingViT}
\defcitealias{su2024roformer}{RoPE}
\defcitealias{lei2016layer}{LayerNorm}
\defcitealias{nichol2021glide}{GLIDE}
\defcitealias{saharia2022photorealistic}{Imagen}
\defcitealias{ho2022imagen}{ImagenVideo}
\defcitealias{betker2023improving}{DALL-E-3}
\defcitealias{zhang2018unreasonable}{LPIPS}
\defcitealias{podell2023sdxl}{SD-XL}
\defcitealias{karras2019style}{StyleGAN}
\defcitealias{raffel2020exploring}{T5-XXL}
\defcitealias{flux}{FLUX.1}
\defcitealias{auraflow}{AuraFlow}
\defcitealias{ni2023conditional}{LFDM}
\defcitealias{zhang2023i2vgen}{I2VGen-XL}
\defcitealias{tian2024reducio}{REDUCIO!}
\defcitealias{siamese}{Siamese Network}
\defcitealias{jolliffe2002principal}{PCA}
\defcitealias{zhang2024pixel}{Pixel-loss}
\defcitealias{kong2024hunyuanvideo}{HunyuanVideo}

\begin{document}

\maketitle
\begin{abstract}
Recent text-to-video diffusion models can generate compelling video sequences, yet they remain silent—missing the semantic, emotional, and atmospheric cues that audio provides. We introduce LTX-2, an open-source foundational model capable of generating high-quality, temporally synchronized audiovisual content in a unified manner.
LTX-2 consists of an asymmetric dual-stream transformer with a 14B-parameter video stream and a 5B-parameter audio stream, coupled through bidirectional audio-video cross-attention layers with temporal positional embeddings and cross-modality AdaLN for shared timestep conditioning.
This architecture enables efficient training and inference of a unified audiovisual model while allocating more capacity for video generation than audio generation. We employ a multilingual text encoder for broader prompt understanding and introduce a modality-aware classifier-free guidance (modality-CFG) mechanism for improved audiovisual alignment and controllability.
Beyond generating speech, LTX-2 produces rich, coherent audio tracks that follow the characters, environment, style, and emotion of each scene—complete with natural background and foley elements.
In our evaluations, the model achieves state-of-the-art audiovisual quality and prompt adherence among open-source systems, while delivering results comparable to proprietary models at a fraction of their computational cost and inference time. All model weights and code are publicly released.
\footnote{\href{https://github.com/Lightricks/LTX-2}{https://github.com/Lightricks/LTX-2}}
\end{abstract}

\section{Introduction}

\begin{figure}[ht]
    \centering
    \includegraphics[width=\linewidth]{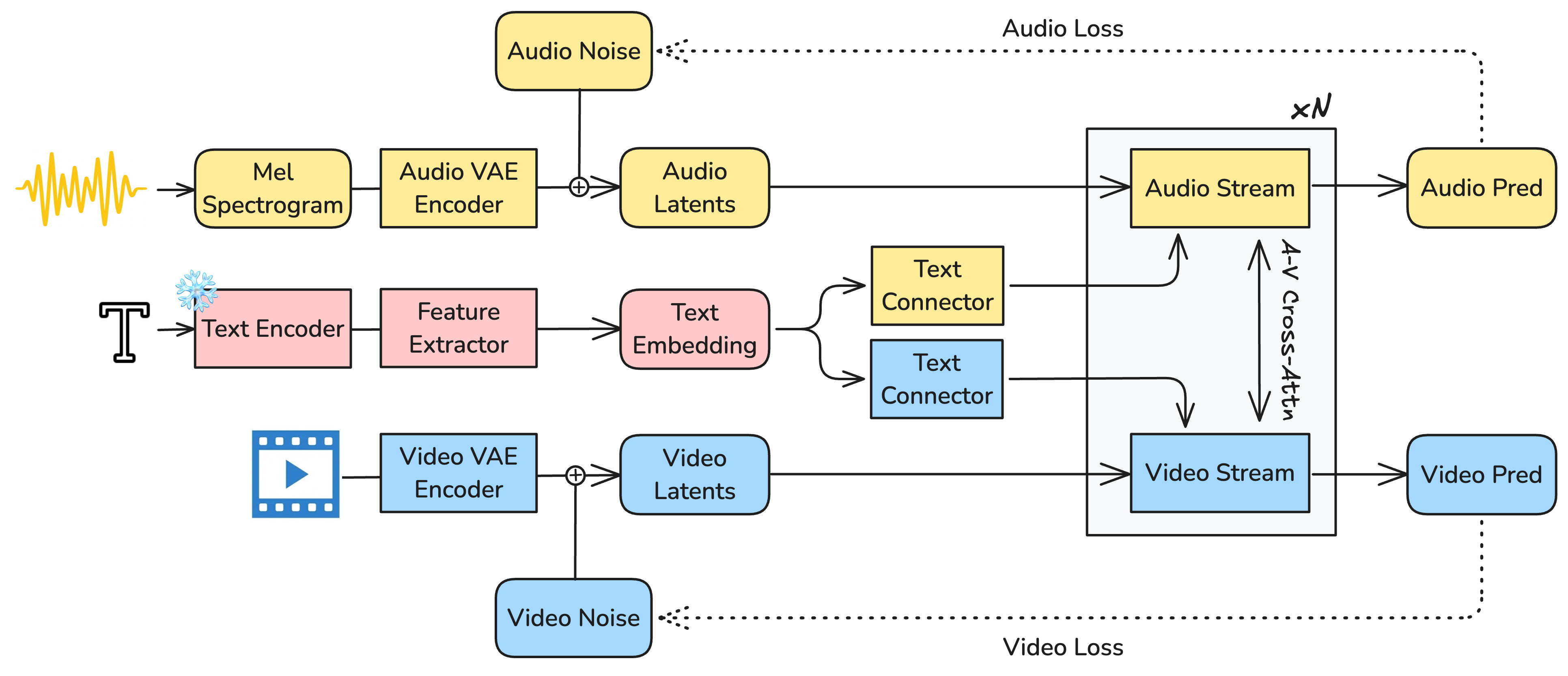}
    \caption{
Overview of the \textit{LTX-2} architecture. Raw video and audio signals are encoded into modality-specific latent tokens via causal VAEs, while text is processed through a refined embedding pipeline. A dual-stream diffusion transformer jointly denoises audio and video latents with bidirectional audiovisual cross-attention and text conditioning, producing synchronized audiovisual outputs.    
    }
    \label{fig:arch-overview}
\end{figure}

Recent text-to-video (T2V) diffusion models have achieved substantial progress, producing videos with striking visual realism, motion consistency, and strong prompt fidelity. Models such as LTX-Video~\cite{hacohen2024ltx}, WAN 2.1~\cite{wang2025wan}, and HunyuanVideo~\cite{kong2024hunyuanvideo} demonstrate how large-scale latent diffusion transformers can translate textual descriptions into temporally coherent and visually expressive video content. Yet these models remain silent: they omit the semantic, emotional, and environmental information conveyed by synchronized sound. As a result, their outputs, while visually stunning, often feel incomplete and offer limited practical utility.

In parallel, text-to-audio (T2A) generation has evolved from task-specific systems toward more general-purpose representations. 
Despite this progress, most text-to-audio models remain specialized for specific domains—such as speech, music, or foley—rather than offering a unified, holistic approach to audio generation.
Consequently, recent attempts to achieve audiovisual generation (T2AV) often rely on decoupled sequential pipelines: generating video (T2V) and then ``filling in'' the audio (V2A), or vice versa. We argue that such decoupled approaches are inherently sub-optimal as they fail to model the full joint distribution of the two modalities. For instance, while lip synchronization is primarily driven by audio, the acoustic environment---such as reverberation or foley---is dictated by the visual context. A unified model is required to capture these bidirectional dependencies.

Achieving a coherent audiovisual experience requires a unified model that jointly captures the generative dependencies between vision and sound. While emerging proprietary systems such as Veo 3~\cite{veo32025} and the concurrent open-source Ovi~\cite{ovi2025} begin to explore this direction, the field still lacks an open, efficient, and high-fidelity text-to-audio+video (T2AV) framework that learns both modalities in an integrated manner.

In developing \textit{LTX-2} as an efficient multimodal foundation model, we prioritize both semantic grounding and computational efficiency. Our architecture builds upon the design principles of \textit{LTX-Video}~\cite{hacohen2024ltx} and its spatiotemporal latent space, while introducing specialized components for high-fidelity audio, multilingual support, and a refined text understanding pipeline---which we found critical for high-quality speech generation and complex prompts. Our design is guided by the following principles:

\textbf{Decoupled Latent Representations.}
Rather than forcing video and audio into a shared latent space, we utilize separate, modality-specific VAEs. This decoupling is fundamental to our approach: it allows us to utilize modality-appropriate positional embeddings (3D for video vs. 1D for audio), independently optimize compression levels for each signal type, and exercise precise control over the model capacity allocated to vision versus sound. Furthermore, this separation natively supports editing workflows, such as generating synchronized audio for an existing video (V2A) or synthesizing video driven by a specific audio track (A2V).

\textbf{Asymmetric dual stream.}
Video and audio possess fundamentally different information densities. We process these modalities through an asymmetric dual-stream transformer architecture. A wide, high-capacity stream handles the complex spatiotemporal dynamics of video, while a narrower, specialized stream processes the 1D temporal nature of audio. This design ensures that computational resources are spent where they are needed most---maintaining high visual fidelity without over-parameterizing the audio pathway.

\textbf{Cross-modal attention.}
To achieve tight temporal alignment, we integrate bidirectional cross-attention layers throughout the model’s depth. By utilizing 1D temporal RoPE during these interactions, the model learns to map visual cues (e.g., the impact of a physical object) to auditory events (e.g., the resulting foley sound) with sub-frame precision. This enables the model to capture complex dependencies like lip-synchronization and environmental acoustics without degrading the unimodal generation quality of either stream.

\textbf{Deep Multilingual Grounding for Speech.}
We found that advanced text understanding is critical not only for global language support but for the phonetic and semantic accuracy of generated speech. By leveraging a high-parameter multilingual text encoder (Gemma 3~\cite{team2025gemma}), a specialized multi-layer feature extraction strategy, and dedicated text processing blocks for multi-token prediction, \textit{LTX-2} achieves a level of prompt adherence that permits highly expressive and accurate speech synthesis. This allows the model to synthesize speech that is not only synchronized with visual lip movement but also natural in its cadence, accent, and emotional tone.

\textbf{Summary of Contributions}

To realize these principles, we introduce several technical innovations. Our key contributions are as follows:

\begin{itemize}

\item \textbf{Efficient Asymmetric Dual-Stream Architecture:}
A transformer-based backbone featuring modality-specific streams linked via bidirectional cross-attention and \textit{cross-modality AdaLN} for shared timestep conditioning.

\item \textbf{Text Processing Blocks with Thinking Tokens:}
A refined text-conditioning module employing multi-token prediction for enhanced prompt understanding and semantic stability.

\item \textbf{Compact Neural Audio Representation:} An efficient causal audio VAE that produces a high-fidelity, 1D latent space optimized for diffusion-based training and inference.

\item \textbf{Modality-Aware Classifier-Free Guidance:} A novel \textit{Bimodal CFG} scheme that allows for independent control over cross-modal guidance scales, significantly improving audiovisual alignment.

\end{itemize}
Through these contributions, \textit{LTX-2} establishes a new open-source foundation for T2AV generation, capable of producing coherent, expressive, and richly detailed content at unprecedented speed.

\section{Related Work}

Diffusion Transformers (DiTs) have emerged as a unifying architecture for large-scale generative modeling. Introduced by Peebles and Xie~\cite{peebles2023scalable}, DiTs replace the traditional U-Net backbone with a transformer operating in latent space, enabling superior scalability and global receptive fields. Subsequent advances in Rectified Flow~\cite{lipman2022flow} have further optimized these models by framing denoising as a continuous flow, reducing sampling steps and improving efficiency. These developments form the architectural foundation for recent advances in multimodal generative models. $\textit{LTX-2}$ builds on this foundation, utilizing an asymmetric DiT backbone optimized for high-throughput multimodal generation.

\subsection{Audio and Video Generation}

\textbf{Text-to-Video Models.}
Recent text-to-video (T2V) models like LTX-Video~\cite{hacohen2024ltx} and WAN 2.1~\cite{wang2025wan} demonstrate the power of DiT architectures trained on massive datasets to produce visually rich, temporally coherent clips. While these models excel at visual realism and motion, they are intrinsically "silent," omitting the auditory dimension that defines immersive content. $\textit{LTX-2}$ extends the efficient spatiotemporal architecture of LTX-Video~\cite{hacohen2024ltx} into the audiovisual domain, introducing a parallel audio stream that maintains visual performance while adding synchronized sound.

\textbf{Decoupled Audio-Visual Synthesis.}
Extensive research has focused on decoupled sequential generation: either Audio-to-Video (A2V)~\cite{luo2023difffoley, gao2025wan} or Video-to-Audio (V2A)~\cite{zhang2024foleycrafter, cheng2025mmaudio, benita2025cafa}. However, these sequential pipelines suffer from an inherent "modality-first" bottleneck. In V2A, the audio model is constrained by a pre-existing video that may lack the necessary visual cues for complex soundscapes. Conversely, A2V models struggle to synthesize realistic environmental foley or background ambiance before the visual scene's details are established. These decoupled approaches fail to capture the recursive nature of audiovisual events, where sight and sound are often markers of the same physical phenomenon.
In contrast, $\textit{LTX-2}$ models the true joint distribution of both modalities, allowing sound and vision to influence each other bidirectionally.

\textbf{Joint Text-to-Audio+Video (T2AV) Models.}
The frontier of T2AV generation involves synthesizing synchronized video and audio from a single text prompt. Proprietary systems like Veo 3~\cite{veo32025} have shown the potential of this joint approach, but their architectures remain closed. Concurrent open-source efforts, such as Ovi~\cite{ovi2025} and BridgeDiT~\cite{bridge2025}, typically duplicate and combine existing T2V and T2A backbones. Such approaches often lead to high computational overhead and limited cross-modal synergy. $\textit{LTX-2}$ differentiates itself by employing a \textit{decoupled yet integrated} dual-stream architecture. By utilizing asymmetric streams and bidirectional cross-attention, we achieve state-of-the-art audiovisual alignment and complete soundscapes (speech, foley, and music) at a fraction of the computational cost of symmetric or proprietary alternatives.

\subsection{Text Conditioning}

The evolution of text-conditioning has moved from training encoders from scratch~\cite{glide2022} to leveraging pretrained encoders such as T5 for scalability~\cite{imagen2022}. Modern approaches often combine frozen encoders with trainable layers, either in parallel to the denoising process~\cite{esser2024scaling, fibo2025} or as an intermediary refinement stage~\cite{kong2024hunyuanvideo}. While T5-like encoders remains a standard choice~\cite{veo32025}, recent models have shifted toward decoder-only LLMs~\cite{sana2024, lumina2024}. However, decoder-only LLMs typically employ causal attention, which can limit the global contextual awareness of the embeddings. To mitigate this, recent work refines these tokens through bidirectional transformer blocks~\cite{kong2024hunyuanvideo}. $\textit{LTX-2}$ adopts this bidirectional refinement but also introduces additional ``thinking'' tokens to the conditioning sequence. This strategy allows the model to aggregate and enrich the text representation before it enters the diffusion cross-attention layers, leading to significantly improved phonetic accuracy in speech and better adherence to complex prompts.

\section{Method}

\textit{LTX-2} is a generative system designed to model the text-conditioned joint distribution of video and audio signals. The model consists of three primary components: (i) modality-specific VAEs that compress raw signals into efficient latent representations; (ii) a refined text embedding pipeline that provides deep semantic and phonetic grounding; and (iii) an asymmetric dual-stream DiT that performs joint denoising through bidirectional cross-modal exchange. See Figure~\ref{fig:arch-overview}.

The audiovisual signal is represented by latent tokens produced by modality-specific VAEs.
\textbf{Video latents} are obtained from a spatiotemporal causal VAE encoder. They are linearly projected to the transformer’s internal width before being processed through a stack of DiT blocks.
\textbf{Audio latents} are derived from mel spectrograms at 16~kHz and encoded by a separate causal audio VAE. These latents are treated as purely temporal sequences and undergo the same processing pipeline as the video tokens.

The processed latents are mapped back to their original dimensionality via learned output projections. The video latents are decoded by the video VAE to produce frames, while the audio latents are decoded by the audio VAE into mel spectrograms, followed by a neural vocoder that reconstructs a 24~kHz waveform.

Text embeddings are processed through dedicated transformer blocks that also predict “thinking tokens”. Both the original and thinking tokens are then fed, via cross-attention, into the dual-stream transformer blocks.

The following sections detail the architectural and conditioning strategies used to achieve high-fidelity synchronized audiovisual generation.

\subsection{Audiovisual Joint Generation}

\begin{figure}[ht]
    \centering
    % Grid removed, keep width
    \begin{overpic}[width=1\linewidth]{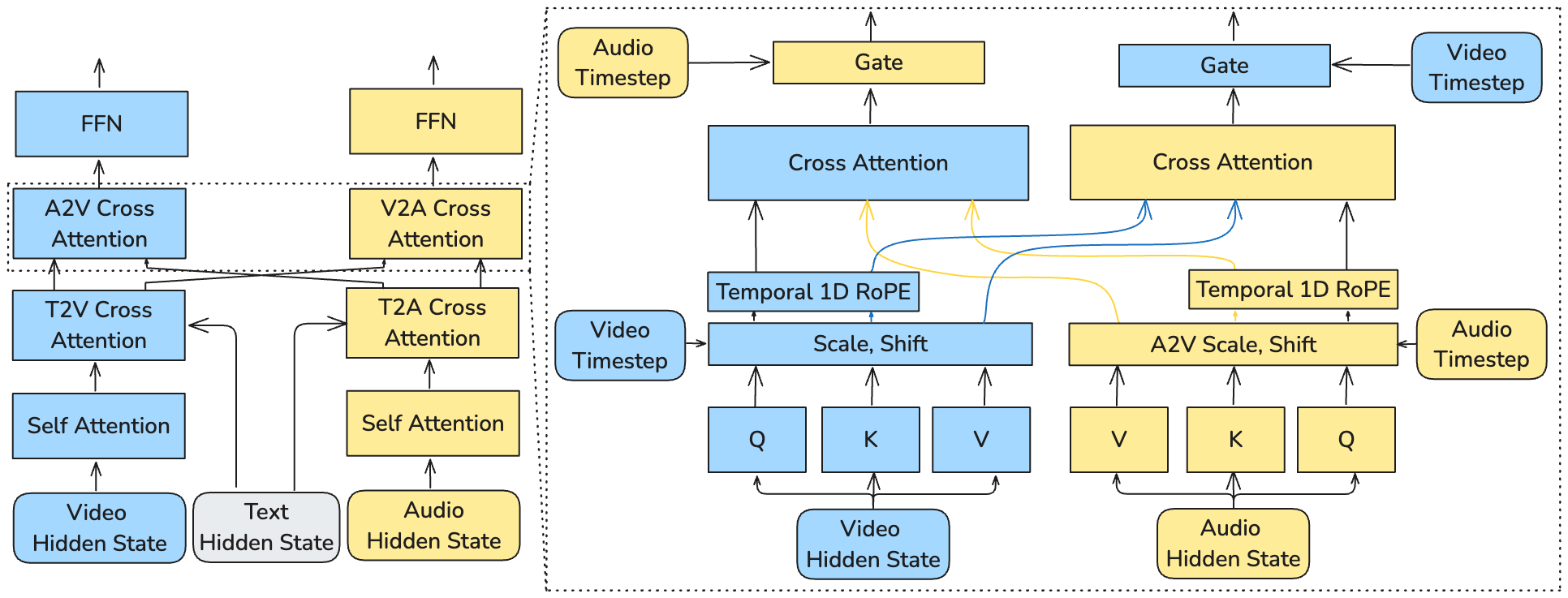}
        
        % (a) - Placing it at X=1, Y=35 (Top Left of first diagram)
        \put(1, -2){\textbf{(a)}} 
        
        % (b) - Placing it at X=34, Y=35 (Top Left of second diagram)
        \put(35, -2){\textbf{(b)}} 
        
    \end{overpic}
    \vspace{0.5em}
    \caption{Proposed architecture. (a) The dual-stream backbone processes video and audio latents in parallel, exchanging information via bidirectional cross-attention layers. (b) Detailed view of the Cross-Attention block, utilizing Temporal 1D RoPE for positional alignment and cross-modality AdaLN for timestep conditioning.}    \label{fig:arch1}
\end{figure}

The core of our system is an asymmetric dual-stream Diffusion Transformer (DiT) architecture. By decoupling the video and audio processing into specialized streams while maintaining a shared depth, we allow each modality to scale according to its own information density. See Figure~\ref{fig:arch1}.

The backbone comprises a high-capacity 14B-parameter video stream and a 5B-parameter audio stream. 
Both streams process latent representations derived from modality-specific causal VAEs.
Each dual-stream block performs four operations sequentially:
Self-Attention within the same modality,
Text Cross-Attention for textual-prompt conditioning,
Audio-Visual Cross-Attention for inter-modal exchange,
and
Feed-Forward Network (FFN) for refinement.
The video stream utilizes 3D Rotary Positional Embeddings (RoPE) to handle spatiotemporal dynamics, while the audio stream uses 1D temporal RoPE. This asymmetry ensures that the majority of parameters are dedicated to the visually complex task of video synthesis, while the audio stream remains efficient.

Throughout the block, RMS normalization layers are interleaved between the main operations, primarily to stabilize activations and maintain consistent scaling across layers. Furthermore, we employ cross-modality AdaLN gates, where the scaling and shift parameters for one modality are conditioned on the hidden states of the other. This allows the model to better synchronize audio and video features, particularly when their diffusion timesteps or temporal resolutions differ.

\subsubsection{Positional Encoding}
The model employs rotary positional embeddings (RoPE) to encode temporal and spatial structure. 
In the \textbf{video stream}, a 3D RoPE injects positional information along spatial and temporal axes $(x, y, t)$, preserving both motion and layout.
In the \textbf{audio stream}, a 1D RoPE is applied along the temporal dimension only.

During audio-visual interaction, only the \textit{temporal} component of RoPE is used for queries and keys, enforcing that cross-modal attention focuses on synchronization in time rather than spatial alignment.

\subsubsection{Audio-Visual Cross-Attention}

\begin{figure}[t]
  \centering
  \includegraphics[width=1    \linewidth]{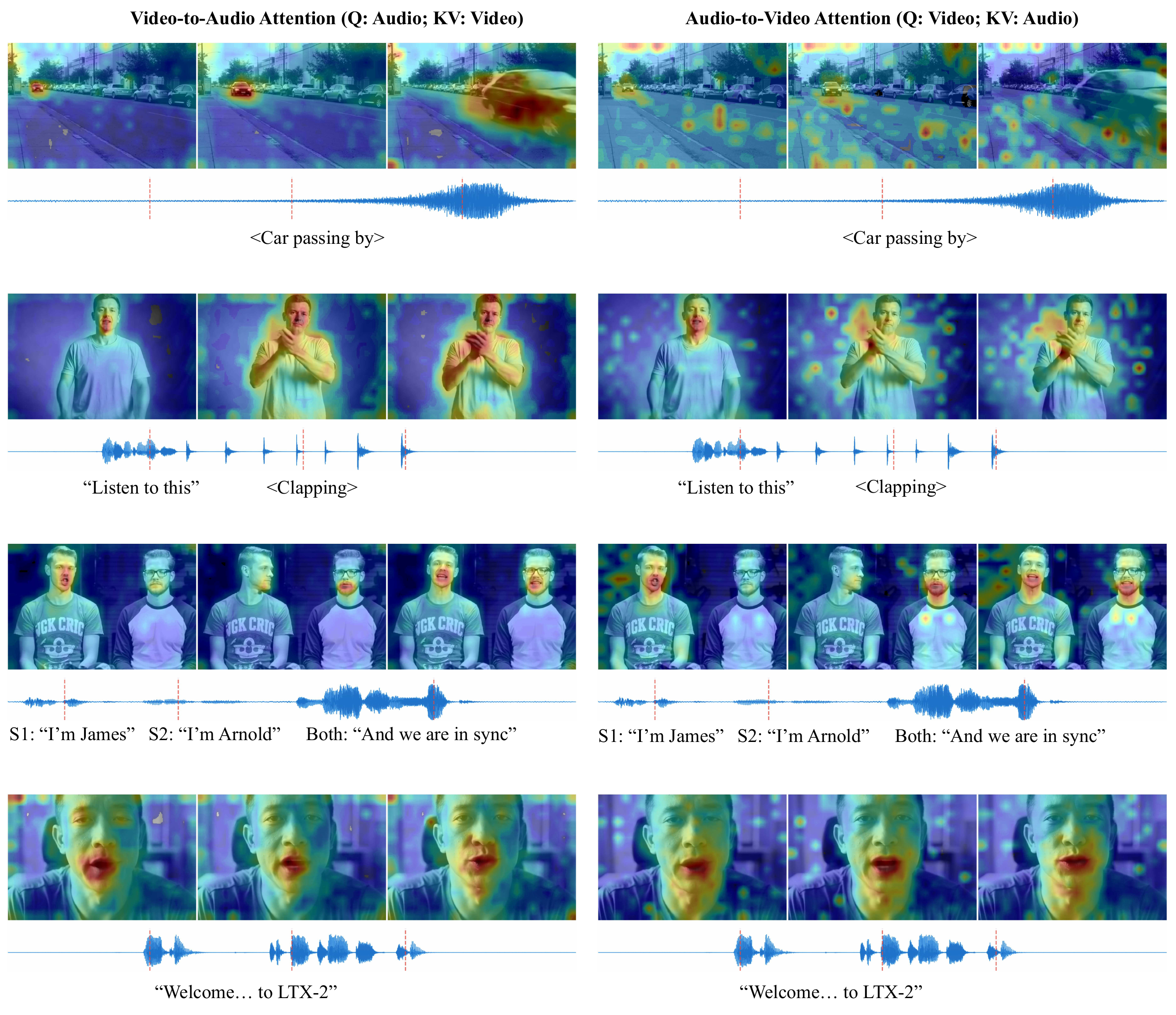}
\caption{\textbf{Visualization of AV cross-attention maps.} The maps are averaged across attention heads and the model layers; V2A and A2V maps correspond to the first and last 1/3 of inference steps, respectively. Red vertical lines on the audio waveform mark the timestamps of the displayed frames. The visualization demonstrates the model's ability to spatially track a moving vehicle, dynamically shift attention from one speaker to another and then to both simultaneously, and focus on the lip region during close-up speech.}  \label{fig:attnmaps}
\end{figure}

At each layer, the audio-visual cross-attention module enables bidirectional information flow between streams.
Both the video and audio hidden states are transformed into queries, keys, and values through learned linear projections that share a common dimensionality.
After these projections, an \textbf{AdaLN modulation} conditioned on the stream’s diffusion timestep scales and shifts the $Q$ and the $(K,V)$ tensors independently, allowing each modality to control how much of its current representation is receptive or exposed to the other.
Temporal rotary embeddings are then applied to $Q$ and $K$, aligning their positions along the shared time axis.
Standard cross-attention is computed between video queries and audio keys/values, and symmetrically in the opposite direction.
The resulting attended representation is passed through an \textbf{additional AdaLN gate}, whose parameters depend on the other modality’s timestep, effectively regulating how much cross-modal information is integrated at that stage of denoising.

Figure~\ref{fig:attnmaps} shows the audio–visual cross-attention maps, illustrating how each modality attends to the most relevant tokens in the other modality.

Additional details about the \textit{LTX-2} architecture are provided in the supplementary material.

\subsection{Deep Text Conditioning and Thinking Tokens}

\begin{figure}[t]
  \centering
  \includegraphics[width=\textwidth]{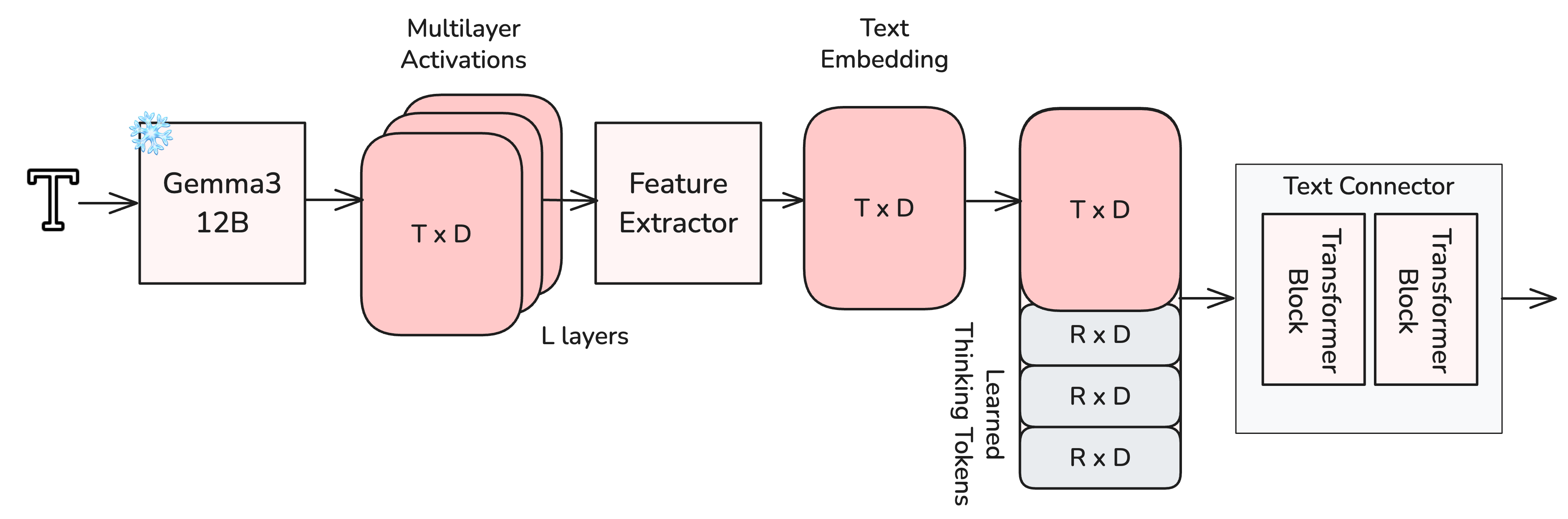}
  \caption{Overview of the Text Understanding pipeline. The text prompt is encoded by Gemma3 and refined through the \textit{Feature Extractor} and \textit{Text Connector} to condition the modality-specific DiT.}
  \label{fig:textconn}
\end{figure}

To support the phonetic precision required for synchronized speech, we move beyond simple global text embeddings. Our conditioning pipeline uses Gemma3-12B~\cite{team2025gemma} as a backbone, refined through two specialized stages (Figure~\ref{fig:textconn}).

For complex conditional generation tasks, relying exclusively on the final-layer embeddings from decoder-only LLMs has been shown to be sub-optimal~\cite{liu2024playground,wang2025comprehensive}. Moreover, Gemma3-12B decoder-only architecture employs causal (unidirectional) attention rather than full bidirectional context modeling. Therefore, we employ the following two methods to compensate for the limitations of causal attention and sub-optimal final-layer embeddings in complex conditional generation.

\subsubsection{Multi-Layer Feature Extractor} Rather than relying on the final causal layer of the LLM, we extract features across all decoder layers. These intermediate representations capture a hierarchy of linguistic meaning—from raw phonetics in the early layers to complex semantics in the later ones. These features are projected into a unified embedding space using a learnable projection matrix W, providing a richer conditioning signal for the diffusion process.

\textbf{Enhanced Feature Representation.}
Studies demonstrate that aggregating information across all decoder layers yields a richer, more comprehensive representation~\cite{fibo2025,liu2024playground,wang2025comprehensive}, providing superior conditioning compared to the output of any single layer. This is primarily because linguistic structure and meaning are distributed across a model's depth~\cite{dar2023analyzing,skean2025layer}.

\textbf{Feature Extractor Design.}
To capitalize on those findings, we designed a dedicated feature extractor that processes the LLM's intermediate layer outputs, which are provided with the shape $[B, T, D, L]$. Where $B$ is the batch size, $T$ the sequence length, $D$ embedding dimension, $L$ number of layers.

The extraction process involves three sequential steps:
\begin{enumerate}
    \item Mean-centered scaling is applied to the intermediate outputs across the sequence and embedding dimensions for each layer.
    \item The scaled output is flattened into a representation of shape $\left[B, T, D \times L\right]$.
    \item This high-dimensional representation is then projected to the target dimension $D$ using a learnable dense projection matrix $\boldsymbol{W}$.
\end{enumerate}

\textbf{Joint Optimization and Freezing.}
The projection matrix $\boldsymbol{W}$ was jointly optimized with the \textit{LTX-2} model during a brief, initial training stage. Crucially, the LLM weights were kept entirely frozen throughout this process.

Optimization was performed using the standard diffusion Mean Squared Error (MSE) loss. This initial joint training yielded an improvement in the model's overall quality. Following this initial stage, the resulting projection matrix $\boldsymbol{W}$ was frozen and maintained for all subsequent training of the \textit{LTX-2} system.

\subsubsection{Text Enhancement with Thinking Tokens}
To enable richer token interactions and contextual mixing before conditioning the diffusion transformer, we introduce a text connector module that jointly processes and refines the text embeddings prior to their integration into the diffusion network.
The text connector consists of transformer blocks with full bidirectional attention that receives embeddings from the feature extractor and refines them before conditioning the diffusion model. It incorporates a learnable set of thinking tokens that are appended to its input tokens, replacing padded positions for improved computational utility.
% Where $E$ denotes the text embeddings, $R$ the learnable thinking tokens to predict, $M$ a binary mask distinguishing valid tokens from padding, $dup$ the duplication operator, and $n=l(E)/l(R)$ the replication ratio:
% \begin{equation}
% M \cdot E + (1 - M) \cdot \operatorname{dup}(R, n)
% \label{eq:registers}
% \end{equation}
%Beyond efficiency, these thinking tokens act as attention sinks and stabilizers, providing the model with dedicated tokens to absorb excess attention and prevent over-mixing of information~\cite{xiao2023efficient, agarwal2025gpt}.
Following recent findings in vision and multimodal transformers~\cite{darcet2023vision, wen2024efficient,pan2025transfer}, the thinking tokens also serve as effective global information carriers, allowing the connector to prepare extra tokens that may carry aggregated contextual information, or missing details that are easier to generate semantically rather than at the visual or audio space.
The resulting sequence, containing both original and thinking tokens, is processed through several transformer blocks and projected via a caption projection layer to form the conditioning input of the diffusion transformer. A separate text connector is assigned to each stream of the transformer, handling the video and audio modalities independently.

The text embedding connectors are trained together with the main audio and video DiT blocks.

\subsection{Audio VAE and Latent Space}
Inspired by the efficient deep latent space introduced in~\cite{hacohen2024ltx}, which demonstrated strong efficiency for video diffusion, we adopt a similarly compact latent representation for audio. Our mel-spectrogram parameterization scheme follows prior work~\cite{liu2023audioldm, audioldm2-2024taslp} on latent diffusion for audio, but extends the audio autoencoder to natively support stereo signals by accepting two-channel mel-spectrograms. Specifically, the input waveform is converted to stereo audio at a 16 kHz sampling rate. We compute a mel-spectrogram for each channel and concatenate the resulting representations along the channel dimension before processing them through the autoencoder. This produces a sequence of latent tokens where each token corresponds to approximately 1/25 seconds of audio and is represented by a 128-dimensional feature vector.

\subsubsection{Vocoder} To reconstruct the final waveform, we utilize a vocoder based on the HiFi-GAN~\cite{kong2020hifi} architecture modified for joint stereo synthesis and upsampling. The generator is conditioned on a two-channel mel-spectrogram (one per stereo channel) computed at 16 kHz and is trained to jointly synthesize a two-channel waveform at a higher sampling rate of 24 kHz. To accommodate the increased complexity of stereo modeling, we double the number of channels in the generator network relative to the original HiFi-GAN V1 design. This increased capacity ensures high-fidelity audio reconstruction and spatial consistency while maintaining the computational efficiency of the 16 kHz latent diffusion process.

\section{Inference}

\subsection{Inference Classifier-free Guidance (CFG)}

During inference, we employ a multimodal extension of Classifier-free Guidance (CFG)~\cite{ho2022classifier} to enhance cross-modal consistency and synchronization, while preserving both video and audio generation quality.  
Our architecture consists of two streams of Transformer blocks, one dedicated to the video stream and one to the audio stream. Each stream is conditioned on textual input as well as on features from the complementary modality through dedicated cross-attention layers.

\begin{figure}[t]
  \centering
  \includegraphics[width=0.6    \linewidth]{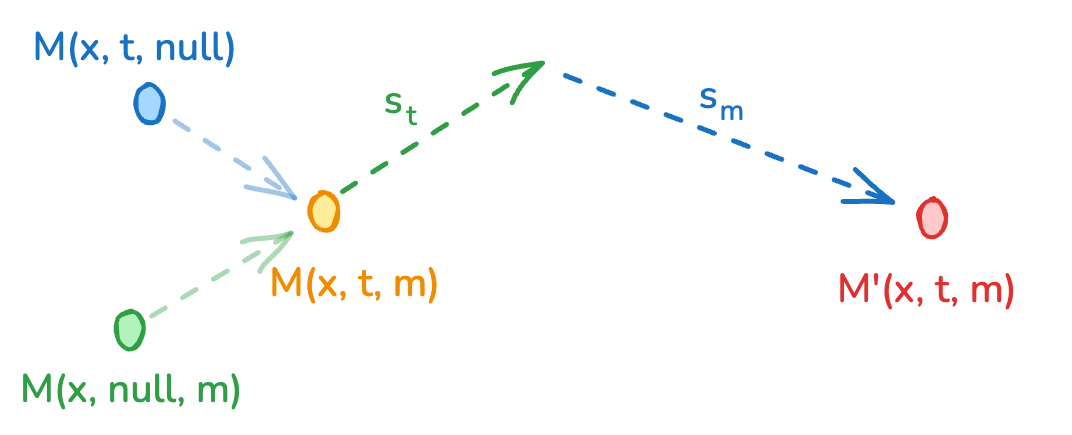}
  % \caption{Overview of the multimodal Classifier-Free Guidance (CFG) mechanism. The model employs text and cross-modal guidance terms to refine generation quality and alignment.}
  \caption{
  Multimodal Classifier-Free Guidance with independent text and cross-modal control. The guided prediction is formed by combining the fully conditioned model output (\textcolor[HTML]{f08c00}{orange}) with two guidance directions: a text guidance term scaled by $s_t$ (\textcolor[HTML]{2f9e44}{green}) and a cross-modal guidance term scaled by $s_m$ (\textcolor[HTML]{1971c2}{blue}). This supports independent control of textual conditioning and inter-modal alignment during inference.
  }
  \label{fig:cross-model-cfg}
\end{figure}

We denote each stream as a model $\mathcal{M}$, which receives as input: (i) $x$: the latent of the current modality (video or audio), (ii) $t$: the text condition, and (iii) $m$: the features of the other modality.\\
To balance the contributions of the textual and cross-modal conditioning, we extend the standard CFG formulation by introducing an additional guidance term for the complementary modality.
For each stream, the guided prediction is computed as:
\[
\hat{\mathcal{M}}(x, t, m) = \mathcal{M}(x, t, m)  \
+ s_t \left( \mathcal{M}(x, t, m) - \mathcal{M}(x, \varnothing, m) \right) 
+ s_m \left( \mathcal{M}(x, t, m) - \mathcal{M}(x, t, \varnothing) \right)
\]
Where $s_t$ controls the strength of the textual guidance, and $s_m$ controls the strength of the cross-modal guidance. As illustrated in Figure~\ref{fig:cross-model-cfg}, this formulation allows independent modulation of text and inter-modal influences during inference. 

Empirically, we observe that increasing $s_m$ promotes mutual information refinement between the modalities. In particular, stronger cross-modal guidance leads to improved temporal synchronization and semantic coherence between generated video and audio, suggesting that this new term effectively aligns the dynamic and contextual information across modalities. For all reported results, we set the guidance weights to $s_t = 3$ and $s_m = 3$ for the video stream, and $s_t = 7$ and $s_m = 3$ for the audio stream.

\subsection{Multi-scale, Multi-tile Inference}

To enable high-resolution synthesis while maintaining the efficiency of our dual-stream architecture, we employ a multi-scale, multi-tile inference strategy. This approach allows LTX-2 to generate Full-HD (1080p) audiovisual content without the memory overhead typically associated with processing high-resolution video latents in a single pass.

\textbf{Base Generation:} Inference begins at a lower resolution, where we generate a "base" latent representation at approximately 0.5 Megapixels (MP). This initial stage establishes the global scene composition, motion dynamics and the foundational audio-visual synchronization.

\textbf{Latent Upscaling:} The base latents are then processed by a dedicated latent upscaler. This module increases the spatial resolution of the video latents while maintaining temporal consistency and auditory alignment, preparing the sequence for high-frequency detail enhancement.

\textbf{Tiled Refinement:} To achieve 1080p fidelity, the upscaled latents are partitioned into overlapping spatial and temporal tiles. Each tile is refined independently using the same foundation model parameters, allowing for the synthesis of intricate visual details—such as skin textures or fine environmental elements—without exceeding GPU memory limits. The tiles are subsequently blended in the latent space to ensure seamless transitions before final VAE decoding.

\section{Training Data}
We used a subset of the same dataset employed in LTX-Video~\cite{hacohen2024ltx}, focusing on video clips that contained significant and informative audio components. This subset provided a balanced distribution of visual and auditory content, allowing us to design captions that fully capture multimodal information relevant to both the image and auditory domains.

\subsection{Captioning}
To generate the high-fidelity textual data required for \textit{LTX-2} training, we developed a new video captioning system capable of describing both the visual and auditory tracks of a clip in exhaustive detail. The system was built to capture every meaningful action, appearance and sound.

Our goal was to create captions that are comprehensive yet factual, describing only what is seen and heard without emotional interpretation. The system captures the full soundscape of each clip, including music, ambient sounds, and precise transcriptions of dialogue with speaker, language, and accent identification. Visual information encompasses camera motion, lighting, and subject behavior. This captioning system provides a comprehensive textual interface between video, audio, and language domains, forming the descriptive foundation for \textit{LTX-2}’s multimodal training corpus.

\section{Experiments}

We evaluate \textit{LTX-2} across three key dimensions: audiovisual quality, visual-only performance via established benchmarks, and computational efficiency. Our results demonstrate that \textit{LTX-2} is not only the highest-performing open-source audiovisual model to date but also delivers unprecedented inference speed.

\subsection{Audiovisual Evaluation}

To assess the quality of joint audiovisual generation, we conducted human preference studies comparing \textit{LTX-2} to both open-source and proprietary state-of-the-art systems. Participants evaluated samples based on visual realism, audio fidelity, and temporal synchronization (e.g., lip-sync and foley accuracy).

Our internal benchmarks indicate that \textit{LTX-2} significantly outperforms open-source alternatives such as Ovi~\cite{ovi2025}. Furthermore, \textit{LTX-2} achieves human preference scores comparable to leading proprietary models, including Veo 3~\cite{veo32025} and Sora 2~\cite{sora22025}. These results establish \textit{LTX-2} as the premier open-source foundation for unified audiovisual synthesis.

\subsection{Video-Only Benchmarks}

While \textit{LTX-2} is a multimodal model, its visual stream maintains top-tier performance on standard video generation tasks. In the \textbf{Artificial Analysis} public rankings (as of November 6th, 2025), \textit{LTX-2} was ranked \textbf{3rd in Image-to-Video} and \textbf{4th in Text-to-Video} generation. Notably, it surpassed proprietary systems such as Sora 2 Pro and large-scale open models like Wan 2.2-14B~\cite{wang2025wan}, demonstrating that our joint training strategy and architectural choices do not compromise visual quality.

\subsection{Inference Performance and Scalability}

The primary advantage of the \textit{LTX-2} architecture is its extreme efficiency. We compared the runtime performance of \textit{LTX-2} (19B parameters, Audio+Video) against Wan 2.2-14B (Video-only) on an NVIDIA H100 GPU. The benchmark was conducted using 121 frames at 720p resolution, with a single-step Euler solver and CFG=1.

\begin{table}[h] \centering \caption{\textbf{Inference Speed.} Comparison of time per diffusion-step on H100 GPU.} \label{tab:performance} \begin{tabular}{@{}l c c c c@{}} \toprule \textbf{Model} & \textbf{Modality} & \textbf{Params} & \textbf{Sec/Step}
\\ \midrule Wan 2.2-14B & Video Only & 14B & 22.30s
\\ \textbf{LTX-2} & \textbf{Audio + Video} & \textbf{19B} & \textbf{1.22s}
\\ \bottomrule \end{tabular} \end{table}

As shown in Table~\ref{tab:performance}, \textit{LTX-2} is approximately 18× faster than Wan 2.2. Due to the optimized latent space mechanism, this performance gap widens further at higher resolutions and longer durations. Furthermore, thanks to its asymmetric design, \textit{LTX-2} is also faster than Ovi~\cite{ovi2025}, which utilizes two 5B transformer streams fine-tuned from Wan 2.2-5B for audiovisual generation.

\textbf{Temporal Scope.} \textit{LTX-2} is capable of generating up to 20 seconds of continuous video with synchronized stereo audio. This exceeds the temporal limits of existing alternatives, including proprietary models like Veo 3 (12s) and Sora 2 (16s), as well as open-source models like Ovi (10s) and Wan 2.5 (10s). This capability makes \textit{LTX-2} uniquely suited for long-form creative content and complex narrative generation.

\section{Limitations}

While \textit{LTX-2} demonstrates strong audiovisual generation capabilities, several limitations remain.
First, performance varies across languages: prompts in languages or dialects underrepresented in the training data may yield less accurate speech synthesis or weaker audio–visual alignment. Second, in multi-speaker scenarios, the model may inconsistently assign spoken content to characters, occasionally confusing which character should speak specific lines.
In terms of temporal scope, generating coherent audiovisual sequences longer than roughly 20 seconds can lead to temporal drift, degraded synchronization, or reduced scene diversity.
Finally, \textit{LTX-2} is a generative diffusion model without explicit reasoning or world-modeling capabilities; deeper narrative coherence, factual grounding, or complex situational understanding depend on external systems such as large language models used to produce the conditioning text.

\section{Social Impact}

Text-to-audio+video generation opens new avenues for creativity, accessibility, and communication. Models like \textit{LTX-2} can enable content creators, educators, and storytellers to produce expressive audiovisual material without requiring specialized equipment or large production teams. The ability to generate synchronized visuals and audio from text has particular promise for low-resource languages and accessibility applications—such as creating inclusive media with speech and sound for visually impaired audiences, or dubbing and localizing educational content across linguistic and cultural boundaries.

At the same time, the technology introduces ethical and societal challenges. Realistic synthetic media carries potential for misuse, including the creation of deceptive or manipulative content. While \textit{LTX-2} is designed for research and creative purposes, responsible use requires clear disclosure of synthetic origin and adherence to content and safety guidelines. Additionally, the model reflects biases present in the data it was trained on, which may manifest in both visual and auditory modalities. Future work should explore methods for bias mitigation, authenticity verification, and improved traceability to ensure safe deployment and positive societal impact.

\section{Conclusion}

We introduced \textit{LTX-2}, an open-source text-to-audio+video (T2AV) foundation model that jointly generates synchronized video and audio from text.
By extending a pretrained 13B video diffusion transformer with a lightweight 3B audio stream connected through bidirectional cross-attention, 1D temporal RoPE, and cross-modality AdaLN conditioning, \textit{LTX-2} achieves efficient multimodal generation without duplicating the visual backbone.
Through modality-aware classifier-free guidance and progressive joint training, the model produces coherent, expressive audiovisual content with natural speech, ambient sound, and foley realism.

Experiments show that \textit{LTX-2} sets a new benchmark for open-source T2AV generation—achieving state-of-the-art audiovisual quality while being the fastest model of its kind.
We hope this work establishes a practical foundation for scalable, accessible audiovisual synthesis and fosters further research in multimodal generative modeling, cross-modal alignment, and controllable sound-aware video generation.

\bibliographystyle{plainnat}
\bibliography{references}

\newpage
\appendix
\onecolumn % Optional: Many conferences prefer single-column for appendices even if the paper is two-column

% Optional: Change figure/table numbering to S1, S2, etc.
\renewcommand{\thefigure}{A\arabic{figure}}
\renewcommand{\thetable}{A\arabic{table}}
\setcounter{figure}{0}
\setcounter{table}{0}

\section{Supplementary Material}

\subsection{Additional Figures}

\begin{figure}[ht]
    \centering
     % Grid removed, keep width
    \begin{overpic}[width=\linewidth]{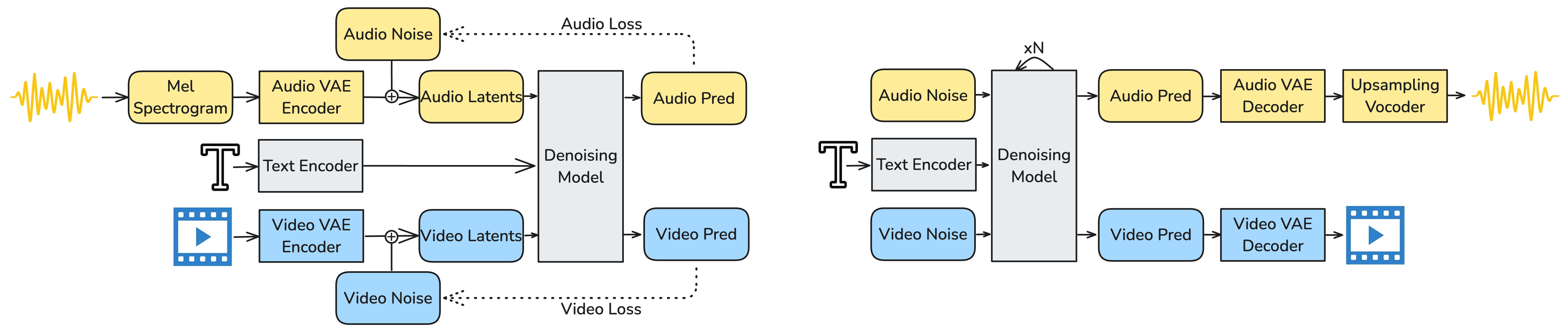}
        
        % (a) - Placing it at X=1, Y=35 (Top Left of first diagram)
        \put(1, 0){\textbf{(a)}} 
        
        % (b) - Placing it at X=34, Y=35 (Toudio-Visual Interaction and Synchronizationp Left of second diagram)
        \put(50, 0){\textbf{(b)}} 
        
    \end{overpic}
    \caption{\textit{LTX-2} training and inference pipelines. (a) Training pipeline: audio and video inputs are encoded into latents, and the model is trained to match their velocity fields using a flow-matching loss. (b) Inference pipeline: starting from noise in the audio and video latent spaces, the model iteratively denoises over $N$ diffusion steps to produce output latents. The VAE decoders and an upsampling vocoder then reconstruct the final waveform and video.}
    \label{fig:pipeline}
\end{figure}

\begin{figure}[h]
    \centering
    \includegraphics[width=0.55\textwidth]{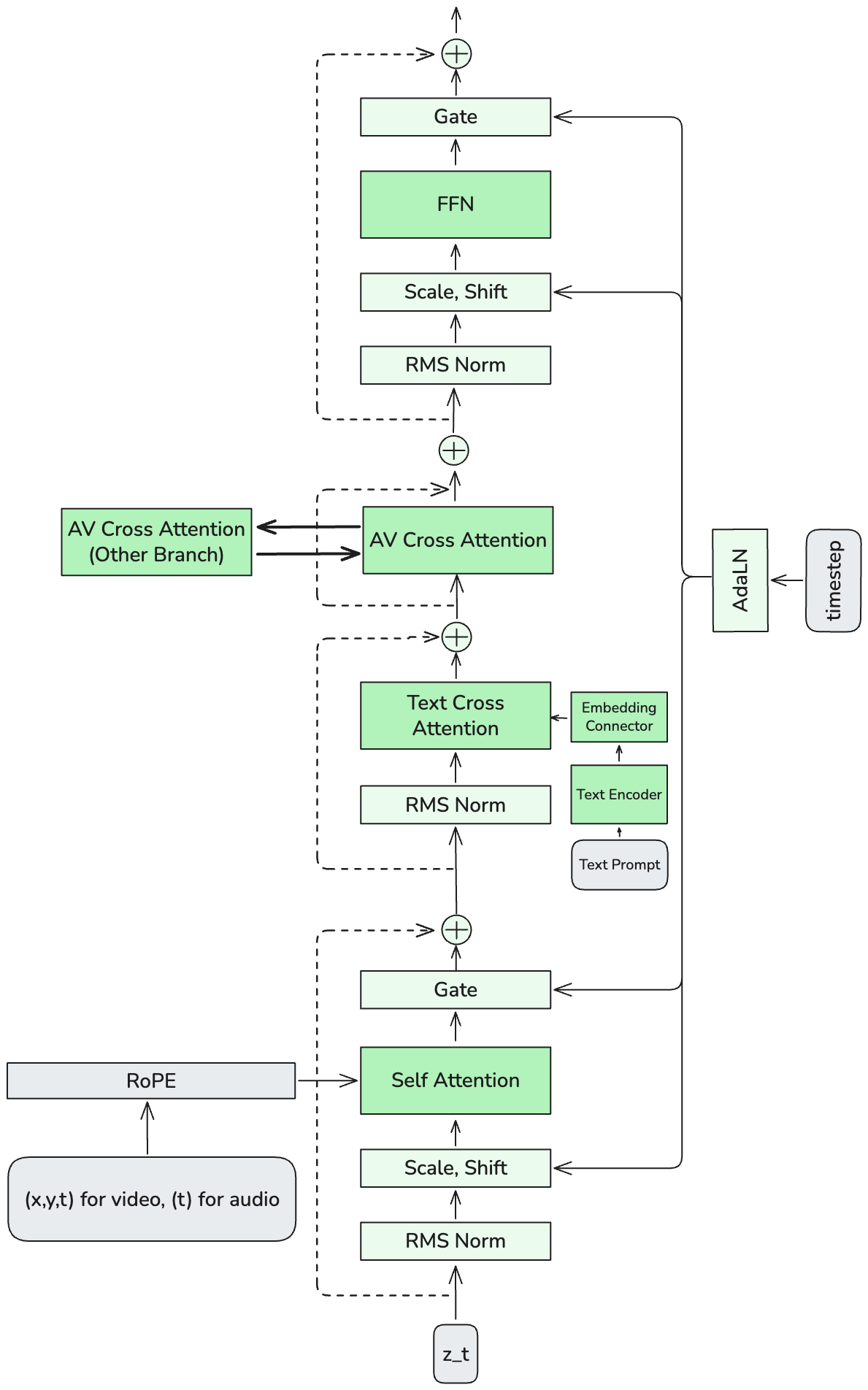}
    \caption{Detailed view of a single stream of the model. The audio and video streams are identical in architecture.}
    \label{fig:single_stream}
\end{figure}
\end{document}